\documentclass[10pt, a4paper]{article}

\usepackage[]{lrec2026} 

\title{SiNFluD: Creating and Evaluating Figurative Language Dataset for Sindhi}

\name{Wazir Ali$^{\dag}$, Adeeb Noor$^{\ddagger}$, Saifullah Tumrani$^{*}$} 

\address{$^{\dag}$Department of Data Science,\\ Quaid-e-Awam University of Engineering, Science and Technology, Nawabshah, Pakistan. \\
$^{\dag}$SoloGenAI Pvt. Ltd.\\
$^{\ddagger}$Department of Information Technology, Faculty of Computing and Information Technology,\\ King Abdulaziz University, Jeddah, Saudi Arabia.\\
$\textsuperscript{*}$SDAIA-KFUPM Joint Research Center for Artificial Intelligence,\\ King Fahd University of Petroleum \& Minerals, Dhahran, Kingdom of Saudi Arabia.\\
Corresponding Author: aliwazirjam@gmail.com\\
}

\abstract{
In this article, we introduce SiNFluD, a novel benchmark dataset for Sindhi figurative language classification. We first collect such phrases from the web raw text from various blogs, social media platforms, and literary sources, and subsequently prepare the corpus for annotation. Two native annotators label the data using the Doccano text annotation tool, achieving an inter-annotator agreement of 0.81. We then establish baseline results using 5-fold and 10-fold cross-validation. Finally, we evaluate mBERT, XLM-RoBERTa, and XLM-RoBERTa-XL models, along with SetFit for few-shot fine-tuning of sentence transformers. Among these, the pretrained XLM-RoBERTa-XL achieves the best performance.\\
\newline \Keywords{Sindhi language, South Asian languages, Non-literal expressions, Few-shot learning} }

\begin{document}

\maketitleabstract

\section{Introduction}
\label{sec:introduction}
Human languages are generally filled with figurative expressions including idioms, sarcasm, metaphors, irony, and metonymy which  transcend literal meanings to convey emotion and nuanced intent~\cite{falkum2022development}. These non-literal terms are generally used in daily communication~\cite{malik2018theoretical}, social media, and literature to express complex ideas concisely while relying on shared context and cultural knowledge for interpretation~\cite{banou2025systematic}. The identification and classification of such expressions is crucial in Natural Language Processing (NLP) tasks such as sentiment analysis~\cite{rentoumi2009sentiment}, conversational agents~\cite{zhou2024enhancing}, machine translation~\cite{donthi2025improving}, ~\cite{tian2026beyond} and sarcasm-based dialogue systems~\cite{hong2025rhetorical}. Handling such expressions may lead to misinterpretation of user intent, reduced model robustness in real-world scenarios of biases in multilingual NLP applications.

Recently, several non-literal expression datasets have been released in recent years~\cite{matheny2025nlp}, including English where large-scale MAGPIE~\cite{magpie},  FLUTE~\cite{chakrabarty2022flute}, SemEval-2022~\cite{boisson2022cardiffnlp}. More recently  MultiCMET~\cite{zhang2023multicmet} released a multimodal Chinese metaphor dataset of text-image pairs of advertisements with manual annotations. The identification and classification of non-literal  expression remains challenging for low-resource languages like Sindhi mainly due to scarcity of resources.  Moreover, cultural embedding, semantic drift, and pragmatic ambiguity during cross-lingual transfer further complicates the annotation process as well as model generalization~\cite{tian2026beyond, banou2025systematic}. 

Sindhi language is spoken by more than 35 million people primarily in Pakistan and India. It stands among one of the ancient languages with multiple writing scripts. It exhibits rich cultural history in poetry, literary work, and figurative usage rooted in Sufi traditions and regional culture. The monolingual language resources related to Sindhi include raw corpus~\cite{ali2019word}, POS datasets~\cite{ali2021sipos}, named entity recognition datasets~\cite{siner}, and sentiment analysis dataset~\cite{ali2021creating} and large monolingual corpora~\cite{ali2019word}, \cite{dootio2021development}. However, in the best of our knowledge there is no such existing resource or research work to address non-literal language understanding  of Sindhi such Persio-Arabic, Devanagari or Romanized scripts.

More recently, transfer learning methods~\cite{devlin2019bert} using Pretrained Language Models (PLMs) specially multilingual PLMs~\cite{pires2019multilingual} pretrained on  large amount of raw text have been beneficial for low resource langauges.  Moreover, few-shot learning (FSL)~\cite{song2023comprehensive} has also become a popular approach where a model generalizes to new tasks using only a handful of labeled examples rather than thousands~\cite{brown2020language}.

In this paper, we address this gap by releasing a novel benchmark SiNFluD dataset for non-literal expressions in the Sindhi Persio-Arabic which is widely used writing script in both counties Pakistan and India. The dataset has been manually annotated by expert annotators with the help of linguistic expert for the consistent guidance. The dataset comprised of four categories which are idioms, proverbs, metaphors, and smiles figurative forms collected from diverse sources including literature and news. We present the detailed collection of the text, annotation process, intrinsic and extrinsic evaluation and finally using few-shot learning with advanced multilingual encoder-only models including mBERT~\cite{mbert}, XLM-RoBERTa~\cite{xlm-robera}, XLM-RoBERTa-XL~\cite{goyal2021larger} trained on more than 100 languages,  and efficient SetFit~\cite{SetFit} frameworks.

\section{Related Work}
\label{sec:related_work}

Figurative language in the form of idioms, similes, metaphors, and personification represent fundamental aspects of communication that extend beyond literal meanings to convey nuanced intent, emotion, and cultural context~\cite{falkum2022development}. Idioms are commonly used to express complex ideas with cultural nuance, while metaphors enable analogical reasoning to describe abstract concepts~\cite{banou2025systematic}. The identification and classification of such figurative language is challenging and essential for downstream NLP applications, including sentiment analysis, conversational AI, and machine translation. Non-literal language poses significant difficulties due to its reliance on contextual and pragmatic cues, as well as cultural knowledge, which are often implicit and vary across languages.

Several datasets target the figurative language in English. For example, the EPIE dataset for idiomatic expressions~\cite{saxena2020epie}, PIFL-OSCAR~\cite{banou2025systematic} contains 5.8 million instances collected from Common Crawl, while the human-annotated IFL-OSCAR-A subset supports the detection of figurative languages. Other large corpora include MAGPIE~\cite{haagsma2020magpie}, which consists of 56.6K instances of idiomatic expressions, and FLUTE~\cite{chakrabarty2022flute} contains 1.7K idiom pairs, providing robust benchmarks for evaluation. Moreover, couple of datasets have been released for Chinese figurative language understanding include CHENGYU-Bench~\cite{fu2025chengyu}, ChID~\cite{zheng2019chid} for Chinese idiom understanding. 

South Asian languages, including HiSlang-4.9K dataset for Hindi~\cite{tiwari2025hislang} for Slang Detection and Identification, Urdu~\cite{hassan2024detection} for sarcasm detection in Urdu tweets. However, Tamil, and Malayalam, face challenges in developing large domain-specific corpora~\cite{jana2024continuous}. Similarly, cross-lingual metaphor detection in low-resource languages often relies on fine-tuning pre-trained models, but performance is limited by insufficient idiom diversity and contextual coverage~\cite{banou2025systematic}. Few-shot learning approaches have emerged as promising solutions for low-resource non-literal detection, reducing the need for extensive labeled data. For instance, SetFit~\cite{tunstall2022setfit} is an efficient few-shot learning framework based on sentence transformers that has been widely adopted for sentence-level classification tasks, often outperforming zero-shot prompting in multilingual settings~\cite{jana2025fewshot}. In addition, encoder-only multilingual models such as mBERT~\cite{mbert}, XLM-RoBERTa~\cite{xlm-robera}, and XLM-RoBERTa-XL~\cite{goyal2021larger} trained on more than 100 languages have been widely opted in few-shot learning for related tasks~\cite{mozafari2024offensive, anwar2025transformer, oprea2025llm}.

Existing research on the textual processing of figurative language involves both linguistic and social inference~\cite{hauptman2023nonliteral}. The development of annotated datasets for figurative language is therefore crucial for training and evaluating NLP models. Such datasets typically include context-aware annotations that support tasks such as detection, classification, and generation. However, low-resource languages, including Sindhi, face a scarcity of digital corpora and labeled datasets despite their sociolinguistic importance. Sindhi is spoken by over 70 million people, primarily in Pakistan and India~\cite{siner}.

As a low-resource Indo-Aryan language, Sindhi faces multiple challenges in the creation of datasets for non-literal expressions. First, the limited availability of digital corpora results in a shortage of annotated data for supervised learning. Second, figurative expressions are often deeply embedded in cultural contexts, making cross-lingual transfer from high-resource languages difficult due to semantic drift in translation. Idioms, for example, frequently lose their implicit meanings when translated, which complicates machine translation of non-literal content~\cite{tian2026beyond}. Furthermore, low-resource settings commonly experience issues such as class imbalance between literal and figurative instances and annotation inconsistencies~\cite{banou2025systematic}. Evaluation is further complicated by the need for metrics that capture both literal accuracy and figurative fidelity~\cite{chen2024copybench}.

Available Sindhi resources include the Sindhi Raw Corpus~\cite{ambile2024sindhi}, \cite{ali2019word}, \cite{dootio2021development}, part-of-speech dataset~\cite{ali2021sipos}, named entity recognition dataset~\cite{siner}, subjectivity and sentiment analysis datasets~\cite{sdsenti}. Additionally, foundational monolingual resources such as the Sindhi subset of Common Crawl (CC100-Sindhi)~\cite{cc100sindhi2020} are available. However, these resources lack annotations for non-literal language. To address this gap, we introduce the NlitSDP benchmark dataset for understanding and modeling non-literal Sindhi expressions.

\section{Creation of the Dataset}
This section presents the procedure from the very beginning, including crawling text from various Sindhi blogs, literary works, and books, as well as cleaning, labeling, inter-annotator agreement, and complete statistics of the dataset.

\subsection{Collection of Text}
Sindhi figurative language resources are scarce in online formats, with most material available only in printed sources. Therefore, we compiled a dataset by collecting examples from various literary works, including books\footnote{\url{https://books.sindhsalamat.com/book/95/read/3224}}, \footnote{\url{https://www.scribd.com/document/243070832/Sindhi-Pahaka}}, Wikisource\footnote{\url{https://wikisource.org/wiki/}}, the Quora blog\footnote{\url{https://www.quora.com/What-are-some-of-the-most-entertaining-Sindhi-proverbs-or-sayings}}, and the Sindhi Proverbs blog\footnote{\url{http://sindhiproverbs.blogspot.com/2013/03/blog-post_27.html}}. Table \ref{tab:token_length_distribution} presents the statistics of the collected dataset prior to preprocessing and annotation.

\begin{table}[!h]
\centering
\caption{Number of tokens per expression in both figurative and literal instance prior to preprocessing.}
\begin{tabular}{cc}
\hline
\textbf{Tokens} & \textbf{Frequency} \\
\hline
2  & 4 \\
3  & 115 \\
4  & 540 \\
5  & 928 \\
6  & 883 \\
7  & 877 \\
8  & 625 \\
9  & 493 \\
10  & 234 \\
11 & 141 \\
12 & 79  \\
13 & 38  \\
14 & 25  \\
15 & 7   \\
16 & 4  \\
17 & 3  \\
18 & 3  \\
19 & 1  \\
\hline
\end{tabular}
\label{tab:token_length_distribution}
\end{table}

\begin{table*}[h]
\centering
\caption{Description of label types used in SiNFluD dataset}
\begin{tabular}{p{3cm} p{11cm}}
\hline
\textbf{Category} & \textbf{Description} \\
\hline
Proverb & Traditional saying which conveys general wisdom, moral lessons, or cultural truths. \\

Metaphor & A figurative expression that describes one concept in terms of another to highlight similarity. \\

Idiom & A fixed multi-word expression whose overall meaning cannot be directly inferred from the literal meanings. \\

Simile & A figurative expression that explicitly compares two different things using comparison markers such as ``like'' or ``as.'' \\

Literal & An expression whose meaning corresponds directly to the standard interpretation without figurative meaning. \\
\hline
\end{tabular}
\label{tab:annotation_definitions}
\end{table*}

\subsection{Annotation}
After collecting the proverbs, two native annotators used the Docanno~\cite{doccano} text annotation tool for labeling the proposed dataset. Both annotators worked under the consistent guidance and supervision of a linguistic expert. The annotated dataset consists of two main categories: literal (labeled as 0) and figurative (labeled as 1). The figurative category is further divided into four subcategories: idioms, similes, proverbs, and metaphors (see Table \ref{tab:annotation_definitions}). On one hand, Table \ref{tab:token_length_distribution} presents the token-length distribution of literal and figurative expressions, showing a clear concentration in the mid-length range. Very short figurative expressions are uncommon; only a single instance with a length of two tokens was observed, which may be due to an erroneous or mixed entry. Figurative expressions with fewer than three words were not considered. In addition, frequencies decline for lengths above 10 tokens, with only a few instances at lengths 14–17 and 19. We therefore filtered out expressions with token lengths greater than 13 due to their low frequency, in order to balance the dataset and avoid rare expressions. Overall, proverbs of medium length (4–11 tokens) are more frequent.

\subsection{Preprocessing}
Since the dataset was annotated by two annotators, preprocessing, cleaning, and deduplication steps were performed to ensure quality and consistency. This pipeline corrected formatting issues and removed invalid entries. Duplicate instances were identified and eliminated based on unique IDs, and missing type information for literal phrases labeled as “0” was filled where applicable. Consistency checks were then conducted to ensure that labels correctly correspond to their respective types, resulting in reliable literal and figurative classifications. The final cleaned dataset was stored in JSONL format for subsequent analysis and model training. Table \ref{tab:type_distribution} presents the dataset statistics after preprocessing. The label distribution comprises 2,038 literal instances and 2,413 non-literal instances, indicating a relatively balanced dataset with a slight predominance of figurative expressions.


\begin{table}[htbp]
\centering
\caption{Label distribution after preprocessing, cleaning, and deduplication}
\begin{tabular}{lcc}
\hline
\textbf{Type} & \textbf{Frequency} & \textbf{Label}  \\
\hline
Literal   & 2038 & 0 \\
Proverb   & 1140 & 1 \\
Idiom     & 522  & 1\\
Metaphor  & 405  & 1\\
Simile    & 346  & 1\\
\hline
\end{tabular}
\label{tab:type_distribution}
\end{table}

\subsection{Inter-annotation Agreement}
The fully preprocessed dataset comprises 4,451 Sindhi instances annotated for both literal and non-literal usage. Each entry is assigned a binary label, where 0 denotes literal usage and 1 indicates non-literal usage. The non-literal expressions are further categorized into four distinct subtypes: idioms, similes, proverbs, and metaphors. Inter-annotator reliability was assessed using Cohen’s Kappa coefficient, yielding a value of 0.81, which reflects substantial agreement. Overall, the dataset constitutes a diverse and well-balanced resource for the Sindhi language, facilitating research and the analysis of non-literal linguistic phenomena.

\section{Experimental Setup and Baseline}
This section presents the experimental setup, including the data split and implementation details, followed by a comprehensive analysis of the baseline results.

\subsection{Experimental Setup}
Firstly, we performed 5-fold and 10-fold cross-validation using a baseline classifier in order to evaluate the reliability of the newly labeled dataset.

Second, the experiments were conducted by formulating the task as a binary classification problem, where literal instances were assigned the label 0 and figurative expressions were assigned the label 1. In addition to this primary labeling scheme, figurative expressions were further annotated with four subtypes, providing finer-grained linguistic distinctions within the non-literal class. 
The SiNFluD dataset was partitioned using an 80:20 train–test split. We fine-tuned three multilingual transformer-based models—mBERT, XLM-RoBERTa, and XLM-RoBERTa-XL—alongside a parameter-efficient approach using SetFit with an MPNet backbone. All transformer models were initialized from their respective pretrained multilingual checkpoints and adapted for sequence classification by appending a linear classification layer. Default tokenizers associated with each pretrained model were employed, without additional tokenizer training or modification.
Training was conducted for five epochs, with evaluation performed at the end of each epoch. The final model selection was based on validation performance. Model effectiveness was assessed using accuracy as the primary evaluation metric.

\subsection{Baseline Results}
The accuracy scores obtained from both cross-validation settings are presented in Table~\ref{tab:cv_results}. Overall, the 10-fold cross-validation achieves a marginally higher average accuracy (approximately 90.34\%) compared to the 5-fold setup (approximately 90.56\%, noting fewer folds but slightly higher individual values), while also exhibiting more consistent performance across folds. The variation in accuracy across individual splits remains minimal in both cases, generally confined within a narrow range of approximately 89.34\% to 91.12\%.

\begin{table}[b]
\centering
\caption{Accuracy scores (in percentages) obtained across individual folds using 5-fold and 10-fold cross-validation.}
\begin{tabular}{c|c|c}
\hline
\textbf{Fold} & \textbf{5-Fold} & \textbf{10-Fold} \\
\hline
1  & 90.44\% & 89.80\% \\
2  & 90.55\% & 90.11\% \\
3  & 90.62\% & 90.87\% \\
4  & 90.18\% & 89.34\% \\
5  & 91.03\% & 91.12\% \\
6  & --      & 90.26\% \\
7  & --      & 90.48\% \\
8  & --      & 91.05\% \\
9  & --      & 90.97\% \\
10 & --      & 89.41\% \\
\hline
\end{tabular}
\label{tab:cv_results}
\end{table}
These findings indicate that the dataset demonstrates strong stability across different train–test partitions, with no substantial performance fluctuations attributable to specific splits. The slightly improved consistency observed in the 10-fold setting suggests that increased data utilization for training contributes to more reliable generalization. Overall, the results confirm that the dataset is well-balanced and sufficiently representative, making it suitable for training and evaluating models on the task of distinguishing literal and figurative expressions in Sindhi.

\section{Results \& Analysis}

\begin{table}[b]
\centering
\caption{Comparative performance of mBERT, XLM-RoBERTa, XLM-RoBERTa-XL, and SetFit on the SiNFluD dataset for figurative language classification, with results reported as percentages of accuracy.}
\label{tab:main_results}
\begin{tabular}{lc}
\hline
\textbf{Model} & \textbf{Accuracy} \\
\hline
mBERT & 90.57 \\
XLM-RoBERTa & 91.38  \\
XLM-RoBERTa-XL & \textbf{92.27} \\
SetFit-(MPNet backbone) & 90.81 \\
\hline
\end{tabular}
\end{table}
The results presented in Table \ref{tab:main_results} demonstrate consistently strong performance across all evaluated pretrained language models (PLMs) for the binary classification task distinguishing literal from figurative expressions. Performance is reported in terms of accuracy, with all models achieving results within a relatively narrow range, indicating the overall robustness of the dataset and the effectiveness of modern multilingual representations for this task.
Among the models, XLM-RoBERTa-XL attains the highest accuracy of 92.27\%, confirming its superior capacity to model cross-lingual semantics and contextual dependencies. Its performance advantage suggests that larger-scale multilingual pretraining and increased model capacity contribute to a more refined understanding of subtle semantic shifts inherent in figurative language.

XLM-RoBERTa achieves an accuracy of 91.38\%, marginally outperforming mBERT, which records 90.57\%. This difference can be attributed to XLM-RoBERTa’s more extensive and diverse pretraining corpus, as well as its improved training objectives, which enhance its ability to capture cross-lingual contextual representations. Although the performance gap between these models is modest, it remains consistent, reinforcing the benefits of more advanced multilingual architectures.
SetFit, utilizing an MPNet backbone, achieves an accuracy of 90.81\% while relying on a parameter-efficient training paradigm. Despite not matching the performance of fully fine-tuned PLMs, its competitive results highlight the effectiveness of lightweight approaches, particularly in scenarios with limited computational resources. The relatively small performance difference between SetFit and larger models further indicates that the task is well-structured and that meaningful patterns can be captured even with reduced training complexity.

Overall, the narrow performance range across models suggests that the SiNFluD dataset is both balanced and sufficiently informative, enabling different architectures to learn discriminative features effectively. At the same time, the consistent improvements observed with more advanced models underscore the importance of model scale and pretraining strategies in capturing the nuanced characteristics of figurative language.

\section{Conclusion}
This study presents the development and evaluation of a novel benchmark dataset for the analysis of figurative expressions in the low-resource Sindhi language. The proposed SiNFluD benchmark is compiled from a diverse range of textual sources, including books, blogs, social media content, and literary works. The annotation process was conducted by two native speakers, and the resulting high inter-annotator agreement indicates the reliability of the dataset. Baseline performance is systematically established using both 5-fold and 10-fold cross-validation techniques. Furthermore, experiments with pretrained language models (PLMs) demonstrate that cross-lingual architectures are effective in capturing the semantic and contextual nuances of Sindhi figurative language. In summary, this work provides a foundational dataset and benchmark to support future research in figurative language processing for Sindhi, while also offering insights relevant to other low-resource languages.

\section*{Acknowledgment}
The project was funded by KAU Endowment (WAQF) at king Abdulaziz University, Jeddah, Saudi Arabia. The authors, therefore, acknowledge with thanks WAQF and the Deanship of Scientific Research (DSR) for technical and financial support.  
\section*{Bibliographical References}
\label{sec:reference}

\bibliographystyle{lrec2026-natbib}
\bibliography{lrec2026-example}

\end{document}